%% file: paper.tex
\documentclass[letterpaper, 10 pt, conference]{ieeeconf}
\IEEEoverridecommandlockouts    
\input{stachnisslab-latex}


\usepackage{url}
\usepackage[capitalize]{cleveref}
\crefname{section}{Sec.}{Secs.}
\Crefname{section}{Section}{Sections}
\Crefname{table}{Table}{Tables}
\crefname{table}{Tab.}{Tabs.}
\usepackage{pifont}
\usepackage{makecell}
\usepackage{adjustbox}
\usepackage{array}
\usepackage{multirow}

\title{\LARGE \bf Deep Reinforcement Learning for Flipper Control \\ of Tracked Robots}

\author{Hainan Pan, Bailiang Chen, Kaihong Huang, Junkai Ren, Xieyuanli Chen, Huimin Lu
  \thanks{All authors are with the College of Intelligence Science and Technology, National University of Defense Technology, China.}%
  \thanks{This work has partially been funded by the National Science Foundation of China under Grant U1913202, U22A2059, and 62203460, Major Project of Natural Science Foundation of Hunan Province (No. 2021JC0004)}%
}

\begin{document}
\maketitle
\thispagestyle{empty}
\pagestyle{empty}

\begin{abstract}
    The autonomous control of flippers plays an important role in enhancing the intelligent operation of tracked robots within complex environments. While existing methods mainly rely on hand-crafted control models, in this paper, we introduce a novel approach that leverages deep reinforcement learning (DRL) techniques for autonomous flipper control in complex terrains. Specifically, we propose a new DRL network named AT-D3QN, which ensures safe and smooth flipper control for tracked robots. It comprises two modules, a feature extraction and fusion module for extracting and integrating robot and environment state features, and a deep Q-Learning control generation module for incorporating expert knowledge to obtain a smooth and efficient control strategy. To train the network, a novel reward function is proposed, considering both learning efficiency and passing smoothness. A simulation environment is constructed using the Pymunk physics engine for training. We then directly apply the trained model to a more realistic Gazebo simulation for quantitative analysis. The consistently high performance of the proposed approach validates its superiority over manual teleoperation.
\end{abstract}

\section{Introduction}
\label{sec:intro}

Tracked robots equipped with four flippers exhibit exceptional terrain traversal capabilities, facilitating efficient navigation through uneven terrain during urban search and rescue (USAR) missions~\cite{kruijff2012ssrr,liu2006cjme}. While adding multiple flippers enhances the traversability of the tracked robots, it also introduces a high degree of control freedom. In complex terrain environments, relying solely on manual control can impose a significant cognitive burden and increase the time required for terrain traversal tasks, potentially impacting the rescue success rate~\cite{kruijff2014star}. Consequently, achieving autonomous control of flippers is paramount in augmenting the intelligent operation of tracked robots in USAR tasks.

Extensive research has focused on addressing the issue of autonomous flipper control. Early studies~\cite{nagatani2008iros,kazunori2007iros,okada2011jfr} primarily centered on analyzing the kinematics of robot with specific terrain structures, incorporating additional constraints and simplifications that limit the extensibility of the traversal model. In contrast, recent research based on Deep Reinforcement Learning (DRL) \cite{mnih2015nature} has allowed researchers to adopt a more practical approach to devise strategies for traversing obstacles.

\begin{figure}[t]
\vspace{0.2cm}
  \centering
  \includegraphics[width=\linewidth]{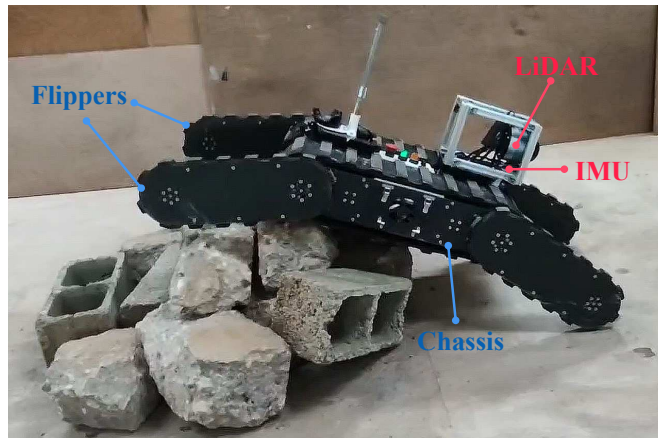}
  \caption{The NuBot-Rescue robot is a tracked robot equipped with four flippers and LiDAR and IMU sensors, showcasing remarkable capabilities in traversing various terrains.}
  \label{fig:robot}
\end{figure}

In this paper, we aim to investigate the DRL for autonomous control of flippers during obstacle traversal by tracked robots, to enhance efficiency and reduce the operational burden on human operators, as shown in~\cref{fig:robot}.
The key contribution of this work lies in the development of a DRL-based flipper autonomous control algorithm that prioritizes the satisfaction of fast-passage and smoothness constraints inherent in the control strategy. 
We design a novel DRL network architecture, dubbed Autonomous Traversal-D3QN (AT-D3QN), consisting of two main modules. The first module fuses the robot and environment states and extracts compact features. We then apply Double Dueling Deep Q-Learing Network (D3QN)~\cite{Mnih2016icml} to provide the proper flipper controls. 
To further enhance the robot's ability to navigate diverse terrain types, we design a novel reward function that considers prior knowledge and the robot's performance in terrain traversal.

After integrating our algorithm, our flipper-based tracked robots acquire the capability to effectively navigate various types of terrain and overcome obstacles. 
To demonstrate the algorithm's generalization capability, the traversal strategy is learned and tested in multiple simulated terrain environments with varying types, sizes, and noise levels. 
The experimental results show that the proposed approach outperforms manual teleoperation, exhibiting improved traffic capacity across different environments.

\section{Related Work}
\label{sec:related}

Autonomous control of articulated tracked robots has been extensively researched for decades. 
Early studies~\cite{li2013amm,nagatani2008iros,kazunori2007iros,okada2011jfr} focused on developing terrain traversal models for the robot in specific terrains, including geometric, kinematic, and even dynamic models. 
Li \etalcite{li2013amm} analyzed the geometric correlation between centroid displacement and the traversal of steps and stairs by a double-flipper crawler robot.
There are also works~\cite{nagatani2008iros,kazunori2007iros,okada2011jfr}developing dynamic models for the sequential locomotion of robots ascending and descending steps and stairs. Such works devise motion strategies for the front and rear flippers based on aligning the robot chassis with the terrain envelope during manual operation as closely as possible. To ensure stability, inappropriate flippers were identified by evaluating the robot's normalized energy stability margin.
These models were analyzed to determine the proper poses of the robot and used for designing the control strategies accordingly. 
However, the analysis of these models was limited to simple and single terrains. When constructing and tackling complex environments, existing works usually fail.

In recent years, machine learning technology has rapidly developed, and researchers have turned their attention to robot control methods based on learning-based techniques.
Paolo~\etalcite{paolo2017arxiv} initially employed a comprehensive end-to-end DRL approach to address the challenge of autonomous flipper control. They utilized Convolutional Neural Network (CNN) to extract depth image features from the robot's front and rear perspectives. These features and the robot's state information were incorporated into the Deep Deterministic Policy Gradient (DDPG) algorithm framework for training purposes. However, the high cost associated with image-based training could have helped to achieve satisfactory outcomes.

Works by Mitriakov~\etalcite{mitriakov2020ssrr,mitriakov2020icfs} optimize the overall stability of the mechanical arm and chassis by incorporating it into the reward function. 
Their methods solely focus on safety constraints, disregarding the potential for enhanced and expedited obstacle traversal. Furthermore, their training solely encompasses stair terrain environments, limiting its applicability to other terrains.
Instead, Zimmermann~\etalcite{zimmermann2014icra} used real terrain traversal data as the expectation, and the feature with the smallest residual in Robot Terrain Interaction (RTI) was extracted by Deep Learning (DL). The Q-learning method was then used to learn the strategy of switching among five predefined flippers from the feature. 
Azayev~\etalcite{azayev2022ral} used data from manual teleoperation, and a state machine network based on imitation learning(IL) was proposed to optimize the unreasonable flipper action switching in~\cite{zimmermann2014icra}. 
Their approach considers the beneficial impact of manual operational expertise on algorithmic control. However, it necessitates substantial data acquisition and learning costs. Furthermore, the algorithm's control effectiveness switches between predefined states and actions, thereby failing to exploit flippers' constructive capabilities during obstacle traversal fully.

From the above research, DRL can better fit the nonlinear contact model between the robot and the terrain by combining DL with Reinforcement Learning (RL). 
This allows for obtaining the action strategy with multiple index constraints, which has great advantages in solving the high-dimensional and complex problem of tracked robots crossing obstacles. 
Nevertheless, attributable to the exorbitant expenses associated with data acquisition and training, achieving substantial advancements in current research poses a major challenge. 
Moreover, numerous studies focusing solely on security constraints lose sight of the beneficial impact of artificial experiential knowledge on algorithms, resulting in a sluggish and incongruous obstacle-traversal effect.

Unlike the above-mentioned approaches, we build a low-cost Pymunk simulation environment for training robots to learn obstacle-crossing strategies. 
To make the robot get better traversal ability in different terrain, a reward function is designed which considers the prior knowledge and smoothness indicators, and an AT-D3QN (Autonomous Traversal-D3QN) flipper autonomous control algorithm suitable for discrete motion space is proposed. The algorithm realizes the migration of the flipper control strategy in a 3D simulation environment and tests it on challenging complex terrain, which can control the articulated tracked robot to cross obstacles quickly and smoothly in real-time.

\section{Our Approach}
\label{sec:main}

\begin{figure}[t]
  \centering
  \includegraphics[width=\linewidth]{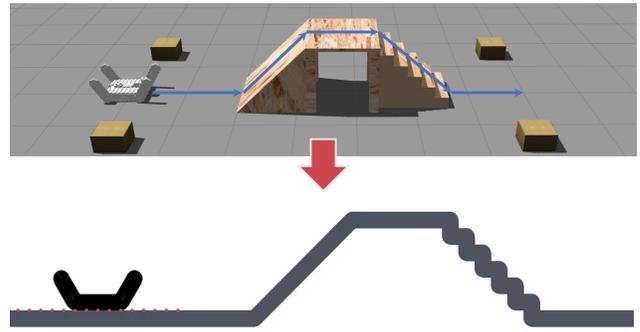}
  \caption{Task Modeling.}
  \label{fig:2D-assume}
\end{figure}

\subsection{Task Modeling and System Overview}
We employed our self-designed NuBot-Rescue robot as the experimental platform, as shown in~\cref{fig:robot}. It incorporates a double-track structure with four flippers, as well as a LiDAR and an IMU sensor for online local mapping and poses estimation. This platform offers advantageous central symmetry properties, and its components, namely the track and flippers, can be controlled independently. In this study, we assume that human operators or path-planning algorithms are responsible for controlling the rotation of the robot tracks, while the developed autonomous flipper algorithm governs the motion of the flippers. 

In real-world scenarios involving traversing complex terrains, it is important to minimize robot instability, such as side-slipping. To achieve this, human operators typically align the robot's forward direction with the undulating terrain of obstacles~\cite{suzuki2014robomech}, employing similar measures for both the left and right flippers.
Building upon this premise, we project the terrain outline and robot shape onto the robot's lateral side. Our approach is particularly suitable for environments with minor left-to-right fluctuations and significant up-and-down fluctuations. An example of the interaction between the robot and the terrain is shown in~\cref{fig:2D-assume}.

In this paper, we use DRL to develop an autonomous control system for the flippers of a tracked robot. Specifically, we formulate the problem as a Markov Decision Processes (MDP) model that leverages the robot's current pose and surrounding terrain data as the state space (see~\cref{sec:ss}) and the front and rear flipper angles as the action space (see~\cref{sec:as}). 
A reward function is established to meet the task's particular requirements (see~\cref{sec:rf}), and subsequently, a novel DRL network is introduced upon the MDP model (see~\cref{sec:an}).

\begin{figure}[t]
  \centering
  \includegraphics[width=\linewidth]{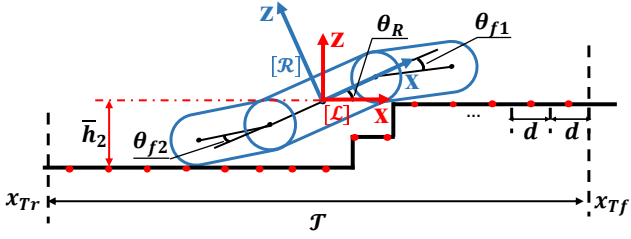}
  \caption{State Space. Includes dynamic interactive information between local terrain and robot.}
  \label{fig:state-space}
  \vspace{-0.3cm}
\end{figure}

\subsection{State Space}
\label{sec:ss}
\textbf{Local Terrain Information $H$:}
The reference coordinate system for local terrain information $H$ is denoted as $[\mathcal{L}]$. 
In this coordinate, the center of the robot chassis serves as the origin, with the X-axis representing the robot's forward direction and the Z-axis indicating the opposite direction of gravity.
To effectively express the RTI, we divide the point set $\mathcal{T}$ consisting of terrain point clouds $l(x_T, z_T)$ in front, back and below the robot into N equally spaced sub-point sets $\mathcal{T}_{i}$, and get N average heights $\overline{h_i}$ as local terrain information representation by downsampling:
\begin{equation} 
	\label{eq:1}
	\begin{aligned}
	H &= \{\overline{h}_i\}=\{\mathop{\mathrm{mean}}\limits_{l(x_T,z_T) \in \mathcal{T}_i}{(z_T)}\},\\
	\mathcal{T} &= \{ \mathcal{T}_i \},\;
	i = 1 \cdots N,\\
	x_T &\in [x_{T_r},x_{T_f}]=[-\frac{N}{2}d,\frac{N}{2}d],
	\end{aligned}
\end{equation}
where $x_T$ and $z_T$ represent the horizontal and vertical coordinates of terrain points in the coordinate system $[\mathcal{L}]$. 
$x_{T_r}$ and $x_{T_f}$ represent the boundaries of the perception domain along the X-axis and cover a range of $[-\frac{N}{2}d,\frac{N}{2}d]$. 
\cref{fig:state-space} depicts the average height $\overline{h}_2$ within sub-point set $\mathcal{T}_2$.

\textbf{Robot State $E$:}
The coordinate system for the rescue robot is defined as a $[\mathcal{R}]$ coordinate system with the center of the robot chassis as the origin, the X axis facing the chassis, and the Z axis perpendicular to the chassis facing upward, as shown in~\cref{fig:state-space} denoted as blue. The robot state $E$ consists of the angle of the robot's front flipper $\theta_{f1}$, the angle of the robot's rear flipper $\theta_{f2}$, and the chassis pitch angle $\theta_{R}$. The angle of the flipper is the X-axis angle between the flipper and the robot $[\mathcal{R}]$ coordinate system, which is positive if the flipper is above the chassis.
The elevation angle of the chassis is the angle between the $[\mathcal{R}]$ coordinate system and the X-axis of the $[\mathcal{L}]$ coordinate system, which is positive if the chassis is above the X-axis of the $[\mathcal{L}]$ coordinate system.

\begin{equation} 
	\label{eq:2}
	\begin{aligned}
	E = \{\theta_{f1}, \theta_{f2}, \theta_{R} \},\\
	\theta_{f1}, \, \theta_{f2}, \theta_{R} \in [-\frac{\pi}{3},\frac{\pi}{3}]
	\end{aligned}
\end{equation}

\subsection{Action Space}
\label{sec:as}
The action of the robot, which governs its movement, is generated as the output of our RL network. The tracked robot possesses the capacity to adjust its posture by rotating its flipper, thereby facilitating efficient traversal over obstacles. 
The action space of the MDP model is designed in the form of discrete angular increments of $\Delta \theta_{f}$ when the flipper rotates, where $ \Delta \theta_{f} = \frac{\pi}{12}$. The front and rear flippers have three motion elements: clockwise rotation of $\Delta \theta_{f}$, counterclockwise rotation of $\Delta \theta_{f}$, and non-rotation. 
Therefore, the motion space is expressed as the combination of nine motions $a$ of the front and rear flipper:
\begin{align}
    \centering
	\label{eq:3}
	A &= \{a_{i,j}\} = \{i \Delta \theta_{f}, j \Delta \theta_{f}\}, \\
	i &=
	\begin{cases}
	-1& \text{front flippers rotate clockwise} \\
	0& \text{front flippers hold on}\\
	1& \text{front flippers rotate counterclockwise}
	\end{cases} \nonumber\\
	j &=
	\begin{cases}
	-1& \text{rear flippers rotate clockwise}\\
	0& \text{rear flippers hold on}\\
	1& \text{rear flippers rotate counterclockwise}
	\end{cases} \nonumber
\end{align} 

\begin{figure}[t]
  \centering
  \includegraphics[width=0.85\linewidth]{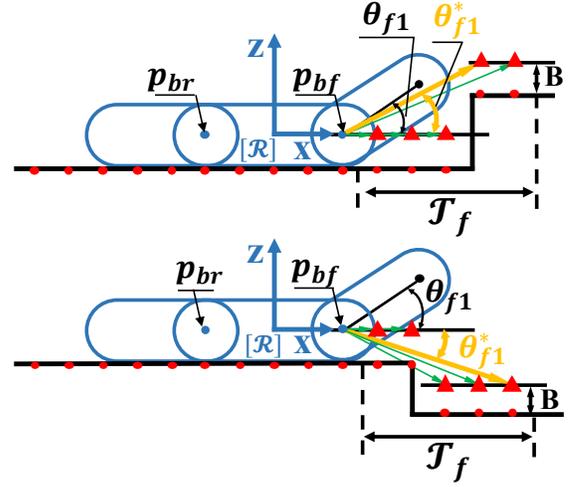}
  \caption{Schematic diagram of reward $R_{flipper}$ calculation.}
  \label{fig:R-flipper}
  \vspace{-0.3cm}
\end{figure}

\begin{figure*}[t]
  \centering
  \includegraphics[width=\linewidth]{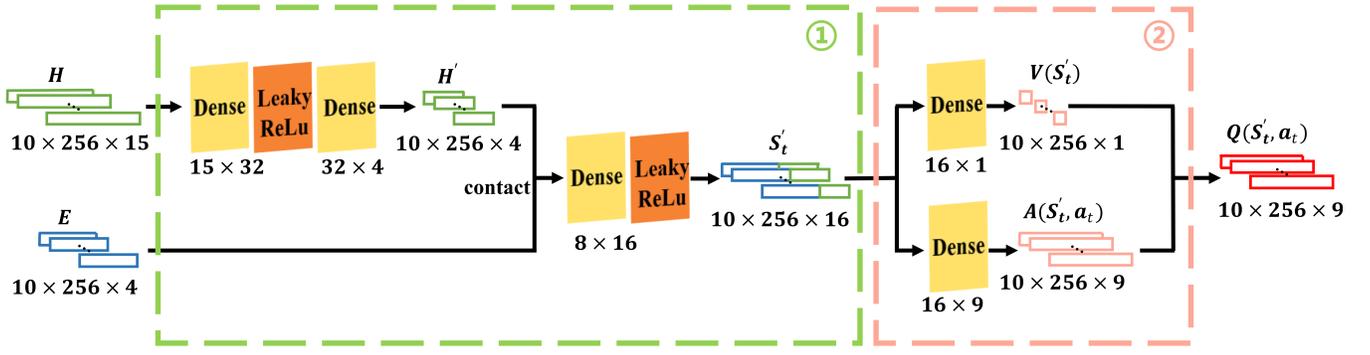}
  \caption{Network structure of our ATD3QN flipper autonomous control algorithm.}
  \label{fig:ATD3QN-net}
  \vspace{-0.3cm}
\end{figure*}

\begin{figure}[t]
  \centering
  \includegraphics[width=\linewidth]{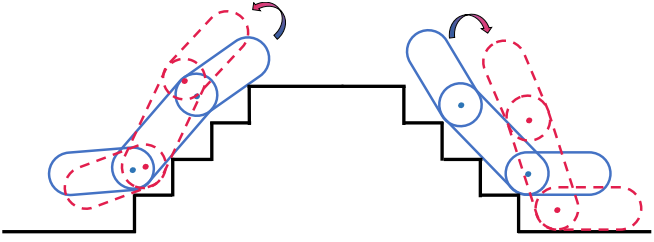}
  \caption{Avoidance of pitching dangerous posture.}
  \label{fig:R-pitch}
  \vspace{-0.3cm}
\end{figure}

\subsection{Reward Function}
\label{sec:rf}
A well-designed reward function is crucial in accomplishing specific tasks for robots \cite{silver2021ai}, as it encourages the learning of efficient control strategies for the flippers. While the front and rear flippers share similar physical characteristics and operational methods during robot terrain traversal, their functional roles are dissimilar. 
Hence, it is necessary to merge prior knowledge and quantitative indicators from expert human operators to devise rewards specific to the front and rear flippers, along with other constraining indicators that enhance the efficiency of the robot learning process.

\textbf{Reward of flipper $R_{flipper}$:}
The front flipper of the robot plays a crucial role in adjusting its posture while enabling the main track to conform to the terrain as much as possible, and it needs to anticipate upcoming obstacles.
We have refined the practical flipper strategy based on the manual teleoperation experience introduced by Okada~\etalcite{okada2011jfr} and designed a motion-based reward function specifically for the front flipper. We denote the reward of front flippers as $R_{flipper}$.

\cref{fig:R-flipper}~(a) and (b) are schematic diagrams of the reward design of the front flipper when the robot goes up and down obstacles, respectively. 
In this section, the hinge point $p_{bf}$ between the front flipper and the chassis is selected as the reference point, which is connected with the expanded terrain point in the point set $\mathcal{T}_f$ as the vector $\overrightarrow{p_{bf}p_{i}}$ (shown as the green vector), and the vector and the robot coordinate system are calculated. 
The one with the largest angle value is selected as the candidate angle of the front flipper (shown by the yellow vector).  
The $R_{flipper}$ is mainly responsible for guiding the robot to change its posture to adapt to the terrain actively. 
A proper reward value is conducive to reducing the little and meaningless action exploration of the robot during training and guiding the robot to explore the reasonable front flipper action more efficiently. 
The absolute difference $\Delta{\theta_{f1}}$ between the robot's front flipper angle $\theta_{f1}$ and the candidate angle $ \ theta _ {f1} $ is taken as the reward index, and $R_{flipper}$ is defined as:

\begin{equation} 
	\label{eq:4}
	\begin{aligned}
	R_{flipper} &=
	\begin{cases}
	-1,& \text{if $\Delta{\theta_{f1}} \textgreater \frac{1}{\lambda_1}$}\\
	-\lambda_{1}{\Delta{\theta_{f1}}},& \text{otherwise}
	\end{cases}\\
    \Delta{\theta_{f1}} &= |\theta_{f1} - (\theta_{f1}^{*} \pm \frac{\pi}{36})|
	\end{aligned},
\end{equation}
where $\lambda_{1}$ is the threshold coefficient of $\Delta{\theta_{f1}}$.

\textbf{Reward of Smoothness $R_{pitch}$:}
Terrain traversal smoothness is an important evaluation standard, and the pitch angle of the robot chassis changes as gently as possible through the cooperation of the rear and front flippers. 
We propose to use the relevant indicators of the robot chassis pitch angle as a reward to optimize the robot's terrain traversal stability, denoted as $R_{pitch}$.

The absolute change of pitch angle is $\Delta|{\theta_{R}}(t)|$ and the average change of pitch angle in $k$ time steps is defined as $\Delta {\theta_{r}^{k}}(t)$:
\begin{equation} 
	\label{eq:5}
	\begin{aligned}
	\Delta|{\theta_{R}}(t)| &= |\theta_{R}(t+1)| - |\theta_{R}(t)| \\
	\Delta{\theta_{R}^{k}}(t) &= \frac{1}{k-1} \sum \limits_{i=t}^{t+k-1} {|\theta_{R}(i+1) - \theta_{R}(i)|}\\
	\end{aligned},
\end{equation}
where $t$ represents the number of steps the robot performs in a single terrain traversal turn as shown in figure ~\cref{fig:R-pitch}, the absolute change of pitch angle $\Delta|{\theta_{R}}(t)|$ reflects that the pitch change trend of the robot chassis is rising ($\Delta|{\theta_{R}}(t)| \textgreater 0$), we limit the situation that the robot is near the overturning boundary, hoping that the pitching trend of the robot will not rise further when it is in high overturning risk; 
The average change of pitch angle within $k$ step $\Delta{\theta_{R}^{k}}(t)$ reflects the stability of the robot's terrain traversal. According to the two related indexes of pitch angle mentioned above, the reward of terrain traversal smoothness is designed as $R_{pitch}$, which is defined as:

\begin{equation} 
	\label{eq:6}
	\begin{aligned}
	R_{pitch} &=
	\begin{cases}
	-1, \quad \text{if $(|\theta_{R}| \textgreater \frac{\pi}{4}$ and $\Delta|{\theta_{R}}(t)| \textgreater 0)$}\\
	-1, \quad \text{if$\Delta{\theta_{R}^{k}}(t) \textgreater \frac{1}{\lambda_2}$} \\
	-\lambda_{2}{\Delta{\theta_{R}^{k}}(t)}, \text{otherwise}\\
	\end{cases},
	\end{aligned}
\end{equation}
where $\lambda_{2}$ is the threshold coefficient of $\Delta{\theta_{R}^{k}}(t)$.

\textbf{Reward of Terminate $R_{end}$:}
In the training process of the RL algorithm, the process from the starting point until the robot meets the end condition is called an terrain traversal turn, and a settlement reward will be given at the end of each turn. 
When the robot reaches the finish line smoothly, and the chassis is close to the ground, it is regarded as a successful obstacle crossing in this round, and a big positive reward is obtained. 
In order to restrain the dangerous or inappropriate behavior of the robot, it is necessary to design negative rewards according to the specific situation of the task scene. Thus, when the robot meets the following conditions, it gets a larger settlement reward and ends the current round:

\begin{equation} 
	\label{eq:7}
	\begin{aligned}
	R_{end} &=
	\begin{cases}
	+R,& \text{reached}\\
	-R,& \text{$|\theta_R| \geq \frac{\pi}{3}$}\\
	-R,& \text{$t \geq t_{max}$}\\
	-R,& \text{stucked}\\
	\end{cases}
	\end{aligned}, 
\end{equation}
among them, $t_{max}$ represents the maximum number of steps the robot performs in a single terrain traversal round, and $R$ represents the value of the settlement reward.

\subsection{Algorithm and Network}
\label{sec:an}

We show our DRL network architecture in~\cref{fig:ATD3QN-net}, which consists of two main blocks. The first block highlighted in green is the robot-terrain interaction feature extraction module, which is responsible for generating the interaction features during the terrain traversal process. 
Terrain data ${H}$~(green vector) and robot information ${E}$~(blue vector) are fed into the network to produce the interaction feature vector ${S^{'}_t}$ via the front-end feature extraction module. 
Due to the relatively small size of the estimated motion vectors, robot state, and terrain information, there is no requirement for designing complex networks similar to those used for images or 3D point clouds~\cite{Kober2013ijrr}. 
The network mainly consists of fully connected layers (yellow) and Leaky ReLu nonlinear activation layers (orange). 
Multilayer Perceptron (MLP) is employed to extract features from terrain information ${H^{'}}$, incorporating them with robot state information ${E}$ to create a new interactive feature vector ${S^{'}_t}$ through single-layer MLP.

The second block highlighted in red is the D3QN module, which combines the advantages of Double DQN \cite{Hasselt2010nips} and Dueling DQN \cite{Wang2016icml}, reflecting the terrain traversal action evaluation score.
It uses the sum of the advantage function $A(S^{'}_t,a_t)$ and the state value function $V(S^{'}_t)$ to estimate the output of the Q network $Q(S^{'}_t,a_t)$. 
The $A(S^{'}_t,a_t)$ measures the importance of a particular action $a_t$ at a given state $S^{'}_t$ compared to the average actions in that state. 
In this way, the Q network is more likely to select actions with higher $A(S^{'}_t,a_t)$.

\section{Experimental Evaluation}
\label{sec:exp}

The proposed method was examined in simulated and real environments across diverse terrain to validate its efficacy. 
This section will address the training procedure, simulation experimentation, and physical experimentation.

\subsection{Training Procedure}

To expedite the training procedure, leveraging the Pymunk physical simulation engine, we established a simulation training environment. 
This setting encompasses a tripartite robot model and a terrain model. 
The simulated robot's parameters align with those of its real-world counterpart. Specifically, the flipper measures 0.536\,m in length, the chassis spans 0.76\,m, and the arc radius is 0.1\,m.

The training scenario stands step and stair, and their respective traversal strategy networks are trained respectively.

\textbf{Step:}
In each episode, the robot traverses one step up or down. Considering the size limit of the robot, the height range of a single step is set to $[-0.4$\,m$, 0.4$\,m$]$, and the height of the step is updated randomly every episode.

\textbf{Stair:}
In each episode, the robot traverses an upward or downward staircase. 
The stair topography consists of several steps of equal size.
The height of each step is 0.2m, the number of steps is $[3, 5]$ and the slope is $33.7^{\circ}$.




\subsection{Qualitative Analysis}

In this section, the feasibility and generalization ability of the algorithm was evaluated through experiments conducted in Gazebo, a three-dimensional simulation environment that closely resembles real-life situations.

\begin{figure}[t]
  \centering
  \includegraphics[width=0.99\linewidth]{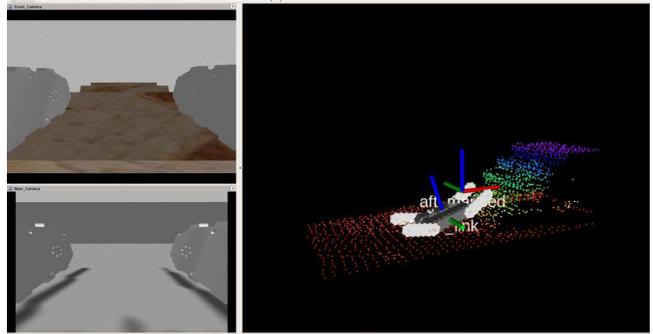}
  \caption{Manual Teleoperation Interactive Interface. It consists of real-time point cloud information, front and rear camera images and robot model pose.}
  \label{fig:manual}
\end{figure}

\textbf{Experimental Conditions.}
The driving mode of the tracked utilizes the simulation control method proposed by Pecka~\etalcite{pecka2017iros}.
The robot is equipped with sensors, including a 32-line LiDAR (Robosense RS-Bpearl) mounted in front and an IMU installed on the chassis. 
The simulation IMU reads the robot's attitude information. 
Terrain information is obtained from the terrain point cloud map generated by the ALOAM algorithm \cite{zhang2014rss}, which combines laser and IMU data. 
The reliable mapping algorithm ensures the accuracy of both terrain and robot pose information, as shown in \cref{fig:manual}. 
Ultimately, the algorithm network takes in 15 downsampled and filtered terrain points along with the real-time robot state as input.

During the experiment, the terrain information was unknown beforehand, and the terrain point cloud map was gradually constructed as the robot moved forward. 
The robot maintained a constant speed of 0.2m/s without adjusting its forward direction. 
The algorithm controlled the entire flipper, with a rotation speed of $25^{\circ}$/s. 

Two types of experimental scenarios were conducted: single step and stair. 
The experiment began and ended 1m from the terrain edge, as depicted in Figure 10. 
In this section, the most challenging single-step terrain with a height of 0.4m was tested, consisting of an upper step and a lower step, spanning a total length of 5m. Additionally, steep stairs, including ascending and descending sections, were tested, covering a total length of 6.5m. 
The specific terrain parameters are provided in \cref{tab:terrain}.

\begin{figure*}[t]
  \centering
  \includegraphics[width=\linewidth]{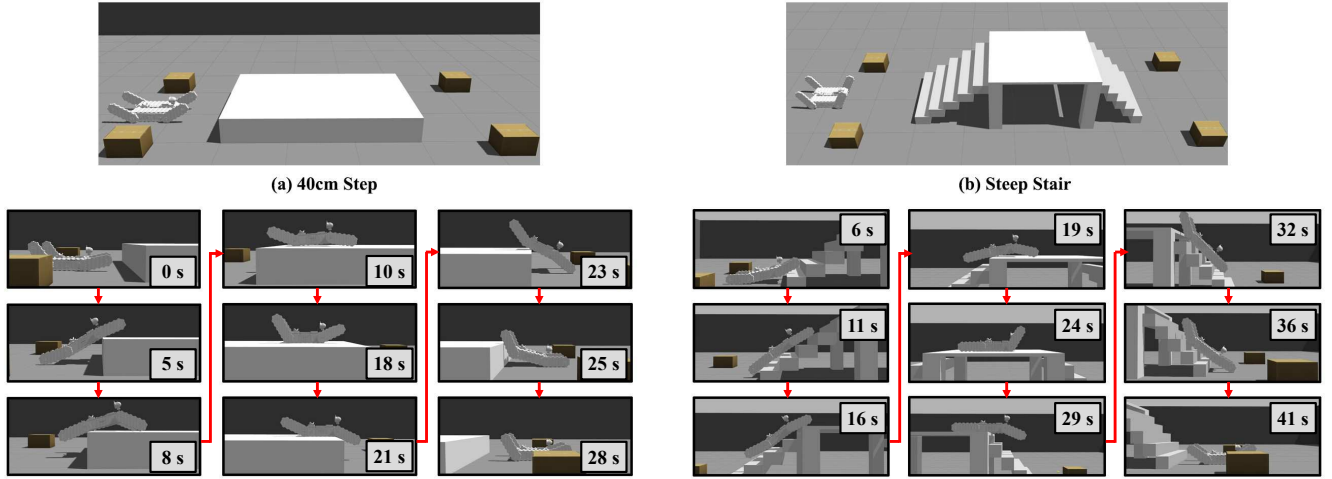}
  \caption{Experimental results of (a)0.4m single-step terrain and (b)steep stair terrain}
  \label{fig:traversal-sim}
\end{figure*}

\begin{table}[t]
  \caption{Parameters of terrains}
  \renewcommand\arraystretch{1.1}
  \setlength{\tabcolsep}{3pt}
  \centering
  \begin{tabular}{lccccc}
    \toprule
    \multirow{2}{*}{Terrain} & \multirow{2}{*}{Type} &Height of  &Length of &Numbers of &Slope\\
            &     &each step  &each step &each step  &    \\
    \midrule
    \multirow{2}{*}{0.4\,m-single-step}  &up/down &0.4\,m  &3\,m  &1 &- \\
     &up &0.2\,m  &0.3\,m  &6 &$33.7^{\circ}$ \\
    \midrule 
    Steep stair &down &0.2\,m  &0.2\,m  &6 &$45^{\circ}$ \\
    \bottomrule
  \end{tabular}
  \label{tab:terrain}
\end{table}

\textbf{Experiment Analysis.}
From the test results shown in \cref{fig:traversal-sim}, the following observations can be made:

(i) For a 0.4m single step, during the time interval of 0-5\,s, the front flipper presses down on the auxiliary robot chassis to climb the step while the rear flipper presses down to maintain a steady pitch rate. During the time interval of 5-8\,s, the robot's center of gravity transitions from the ground to the step. The front flipper depresses in advance to alleviate the impact when reaching the step. From 8\,s to 18\,s, the robot retracts its front and rear flippers and moves forward on a flat surface. From 18\,s to 21\,s, the robot perceives the downward trend of the front step and lowers the front flipper for probing. During the time interval of 21\,s to 25\,s, the front flipper supports reducing the pitch angle change when descending the step, and just before reaching the ground, the front flipper lifts to make contact with the ground. Finally, from 25-28\,s, the robot successfully traverses the stairs and reaches the destination.

(ii) For steep stairs, during the time interval of 0-6\,s, the front flipper actively adapts to the upward trend of the stairs, pressing down to lift the front side of the chassis, allowing the main track to climb along the stairs. From 6-16\,s, the robot maintains a gentle pitch angle that conforms to the terrain slope while climbing. During the time interval of 16-19\,s, the center of gravity of the robot transitioning from the stairs to the platform, and the robot experiences significant oscillations due to the influence of gravity. The front flipper depresses to cushion the impact of reaching the platform. From 24-29\,s, due to the steepness of a 45° slope, while the front flipper probes downward, the rear flipper assists the chassis in adapting to the pitch angle. From 32-41\,s, the robot is about to reach flat ground, and the rear flipper maintains contact with the terrain to reduce vibrations while the front flipper gradually lifts, allowing the main track to make contact with the ground.

The output of the flipper motion generated by the algorithm allows the robot to traverse challenging steps and stairs with varying slopes smoothly. 
The algorithm successfully transfers from a 2D to a 3D simulation environment, demonstrating the adaptability and generalization capability of the traversal strategy obtained by our algorithm for terrains with different parameters.

\subsection{Quantitative Analysis}

\begin{table}[t]
  \caption{Indicator comparison between manual and our work}
  \renewcommand\arraystretch{1.1}
  \setlength{\tabcolsep}{2pt}
  \centering
  \begin{tabular}{l|cc|cc}
    \toprule
    \multirow{2}{*}{Metrics} & \multicolumn{2}{c|}{Manual Teleoperation} & \multicolumn{2}{c}{ATD3QN (Ours)} \\
              &0.4\,m single-step & Steep stair  & 0.4\,m single-step & Steep stair    \\
    \midrule
    $t_{cost}$ [s]$\downarrow$  &46.17 & 76.53 &\textbf{28.07}  &\textbf{41.21}\\
    $\hat{\theta}_R$ [rad]$\downarrow$ &4.02 & 6.30 &\textbf{3.43} & \textbf{4.53}\\
    $\hat{\theta}_R^{range}$ [rad]$\downarrow$ &2.92 & 3.88     &\textbf{1.11} & \textbf{2.60}\\
    \bottomrule
  \end{tabular}
  \label{tab:result}
\end{table}

In this section, the algorithm's control performance is compared quantitatively with that of human operators to demonstrate the practicality and superiority of the algorithm. 

Typically, human operators remotely control the robot using image information from the front and rear cameras and the robot's attitude information. This paper employs the Gazebo simulation environment, utilizing camera and IMU plugins to simulate the operator's perspective. The operator controls the robot's movement through a handle, as depicted in \cref{fig:manual}(a)(b), which shows real-time images from the front and rear cameras on the robot chassis. \cref{fig:manual}(c) also displays the robot's posture, flipper angle, and real-time drawing effect.

In both the 0.4m single-step scenario (\cref{fig:traversal-sim}(a)) and the steep stairs scenario (\cref{fig:traversal-sim}(b)), traversal data were collected for five professional operators, each repeating the experiment three times. The flipper control algorithm was also repeated five times in these two scenarios.

\begin{figure}[t]
  \centering
  \includegraphics[width=0.95\linewidth]{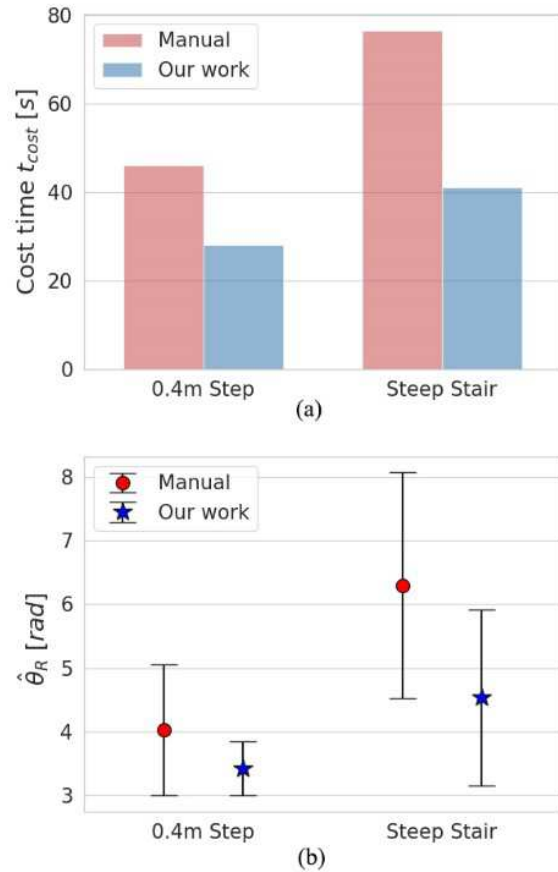}
  \caption{The comparison results between our algorithm and manual teleoperation in two task terrains.}
  \label{fig:result-sim}
\end{figure}

\cref{fig:result-sim} presents the comparison results between the algorithm and the human operator in two different task scenarios. The indicators analyzed include the time-consuming factor $t_{cost}$ for a single mission and the sum of the absolute value of the robot's pitch angle change rate $\hat{\theta}_R$ for a single mission, measured in seconds (s) and radians (rad), respectively. \cref{fig:result-sim}(a) and (b) display the average indicator values and their variations under different control methods in the two task scenarios.
\cref{tab:result} provides specific values for the indicators, including $t_{cost}$, $\hat{\theta}_R$, and the difference $\hat{\theta}_R^{range}$ between the maximum and minimum values of $\hat{\theta}_R$.

By examining the $t_{cost}$ index comparison, it becomes evident that the algorithm proposed in this paper exhibits clear advantages over the human manual operation. Human operators often encounter situations where they need to stop and wait when crossing unknown terrain or adopt conservative and cautious actions when facing challenging terrains.

The time consumption of human operation differs significantly from that of the algorithm, making it inappropriate to rely solely on the average value of the robot's pitch angle change rate $|\dot{\theta}_R|$ to evaluate its performance in single-pass obstacle-crossing tasks. To comprehensively assess task time consumption and pitch angle change rate, this paper introduces a new index $\hat{\theta}_R$: the integral of the absolute value of the robot's pitch angle change rate $|\dot{\theta}_R|$ over the mission time $t_{cost}$. \cref{eq:8} demonstrates the calculation method, where t represents the total number of steps in a single mission:

\begin{equation} 
\label{eq:8}
\begin{aligned}
\hat{\theta}_R = \int_0^{t_{cost}} |\dot{\theta}_R| dt = \sum_{t=1}^{T-1}|\theta_{R}(t+1)-\theta_{R}(t)|
\end{aligned}
\end{equation}

Through index $\hat{\theta}_R$ comparison, it is evident that the average value of the algorithm proposed in this paper is smaller than that of human manual operation in two task terrains: 0.4m single step and steep stairs. This indicates that the algorithm effectively maintains stability throughout the entire obstacle-crossing process. Additionally, when considering the maximum and minimum values $\hat{\theta}_R^{range}$, the algorithm exhibits a narrower range than human operation, highlighting its superior stability in controlling the robot during traversal.

In conclusion, qualitative and quantitative analysis of the experimental results demonstrates that the flipper autonomous control algorithm presented in this paper successfully handles challenging single-step and steep stairs scenarios. The algorithm outperforms human manual operation regarding performance indexes, making it a feasible and adaptable solution.

\section{Conclusion}
\label{sec:conclusion}


This paper proposes a flipper control algorithm AT-D3QN for a tracked robot based on DRL to address the obstacle-crossing challenges faced by flipper-based tracked robots. Our algorithm utilizes the robot's state information and local terrain data and learns to flipper control actions through a reward function that combines obstacle-crossing experience and optimization. This enables the robot to navigate and overcome obstacles smoothly.
The performance of our AT-D3QN is evaluated in two simulated challenging terrain environments using Gazebo. The results demonstrate that the algorithm effectively adjusts the flipper angle of the tracked robot, facilitating smooth obstacle traversal. AT-D3QN also exhibits a certain level of generalization. Furthermore, a quantitative comparison is conducted between the algorithm and manual operation. The results indicate that the flipper control strategy learned by the algorithm outperforms manual operation in two important metrics: overall stability of pitch angle change and obstacle-crossing time. This validates the algorithm's capability to enhance efficiency and safety in robot obstacle-crossing tasks.


\bibliographystyle{plain_abbrv}

\bibliography{glorified,new}

\end{document}

%% file: stachnisslab-latex.tex
\usepackage{graphics}           
\usepackage{times}              
\usepackage{amsmath}            
\usepackage{amssymb}            
\usepackage{graphicx}
\usepackage{algorithm}
\usepackage[noend]{algpseudocode}
\usepackage{booktabs}
\usepackage{color}
\definecolor{instructioncolor}{rgb}{.5,.5,.5}

\usepackage[font=small]{caption}

\def\eqref#1{Eq.~(\ref{#1})}

\makeatletter
\usepackage{xspace}
\DeclareRobustCommand\onedot{\futurelet\@let@token\@onedot}
\def\@onedot{\ifx\@let@token.\else.\null\fi\xspace}

\def\etal{{et al}\onedot}
\makeatother

\def\etalcite#1{\etal~\cite{#1}}

\usepackage{array}
\newcolumntype{L}[1]{>{\raggedright\let\newline\\\arraybackslash\hspace{0pt}}m{#1}}
\newcolumntype{C}[1]{>{\centering\let\newline\\\arraybackslash\hspace{0pt}}m{#1}}
\newcolumntype{R}[1]{>{\raggedleft\let\newline\\\arraybackslash\hspace{0pt}}m{#1}}










%% file: paper.bbl
\begin{thebibliography}{10}

\bibitem{azayev2022ral}
T.~Azayev and K.~Zimmermann.
\newblock Autonomous state-based flipper control for articulated tracked robots
  in urban environments.
\newblock {\em IEEE Robotics and Automation Letters (RA-L)}, 7(3):7794--7801,
  2022.

\bibitem{Hasselt2010nips}
H.~Hasselt.
\newblock Double q-learning.
\newblock In {\em Proc.~of the Advances in Neural Information Processing
  Systems (NIPS)}, volume~23, 2010.

\bibitem{Kober2013ijrr}
J.~Kober, J.A. Bagnell, and J.~Peters.
\newblock Reinforcement learning in robotics: A survey.
\newblock {\em Intl.~Journal~of Robotics Research (IJRR)}, 32(11):1238--1274,
  2013.

\bibitem{kruijff2012ssrr}
G.J.M. Kruijff, F.~Pirri, M.~Gianni, P.~Papadakis, M.~Pizzoli, A.~Sinha,
  V.~Tretyakov, T.~Linder, E.~Pianese, S.~Corrao, F.~Priori, S.~Febrini, and
  S.~Angeletti.
\newblock Rescue robots at earthquake-hit mirandola, italy: A field report.
\newblock In {\em Proc.~of the Intl.~Symposium~on Safety, Security, and Rescue
  Robotics (SSRR)}, pages 1--8, 2012.

\bibitem{kruijff2014star}
G.~Kruijff, M.~Janiek, S.~Keshavdas, B.~Larochelle, H.~Zender, N.~Smets,
  T.~Mioch, M.~Neerincx, J.~Diggelen, F.~Colas, M.~Liu, F.~Pomerleau,
  R.~Siegwart, V.~Hlava, T.~Svoboda, T.~Petriek, M.~Reinstein, K.~Zimmermann,
  F.~Pirri, M.~Gianni, P.~Papadakis, A.~Sinha, P.~Balmer, N.~Tomatis, R.~Worst,
  T.~Linder, H.~Surmann, V.~Tretyakov, S.~Corrao, S.~Pratzler-Wanczura, and
  M.~Sulk.
\newblock Experience in system design for human-robot teaming in urban search
  and rescue.
\newblock {\em STAR Springer Tracts in Advanced Robotics}, 92:111--125, 2014.

\bibitem{li2013amm}
Y.W. Li, S.R. Ge, X.~Wang, and H.B. Wang.
\newblock Steps and stairs-climbing capability analysis of six-tracks robot
  with four swing arms.
\newblock {\em Applied Mechanics and Materials}, 397:1459--1468, 2013.

\bibitem{liu2006cjme}
J.~Liu, Y.~Wang, B.~Li, and S.~Ma.
\newblock Current research, key performances and future development of search
  and rescue robots.
\newblock {\em Jixie Gongcheng Xuebao/Chinese Journal of Mechanical
  Engineering}, 42(12):1--12, 2006.

\bibitem{mitriakov2020ssrr}
A.~Mitriakov, P.~Papadakis, S.~Mai~Nguyen, and S.~Garlatti.
\newblock Staircase negotiation learning for articulated tracked robots with
  varying degrees of freedom.
\newblock In {\em Proc.~of the Intl.~Symposium~on Safety, Security, and Rescue
  Robotics (SSRR)}, volume 2020, pages 394--400, 2020.

\bibitem{mitriakov2020icfs}
A.~Mitriakov, P.~Papadakis, S.M. Nguyen, and S.~Garlatti.
\newblock Staircase traversal via reinforcement learning for active
  reconfiguration of assistive robots.
\newblock In {\em Proc.~of the IEEE Intl.~Conf.~on Fuzzy Systems}, volume 2020,
  pages 1--8, 2020.

\bibitem{Mnih2016icml}
V.~Mnih, A.P. Badia, L.~Mirza, A.~Graves, T.~Harley, T.P. Lillicrap, D.~Silver,
  and K.~Kavukcuoglu.
\newblock Asynchronous methods for deep reinforcement learning.
\newblock In {\em Proc.~of the Int.~Conf.~on Machine Learning (ICML)},
  volume~4, pages 2850--2869, 2016.

\bibitem{mnih2015nature}
V.~Mnih, K.~Kavukcuoglu, D.~Silver, A.A. Rusu, J.~Veness, M.G. Bellemare,
  A.~Graves, M.~Riedmiller, A.K. Fidjeland, G.~Ostrovski, et~al.
\newblock Human-level control through deep reinforcement learning.
\newblock {\em Nature}, 518(7540):529--533, 2015.

\bibitem{nagatani2008iros}
K.~Nagatani, A.~Yamasaki, K.~Yoshida, T.~Yoshida, and E.~Koyanagi.
\newblock Semi-autonomous traversal on uneven terrain for a tracked vehicle
  using autonomous control of active flippers.
\newblock In {\em Proc.~of the IEEE/RSJ Intl.~Conf.~on Intelligent Robots and
  Systems (IROS)}, pages 2667--2672, 2008.

\bibitem{kazunori2007iros}
K.~Ohno, S.~Morimura, S.~Tadokoro, E.~Koyanagi, and T.~Yoshida.
\newblock Semi-autonomous control system of rescue crawler robot having
  flippers for getting over unknown-steps.
\newblock In {\em Proc.~of the IEEE/RSJ Intl.~Conf.~on Intelligent Robots and
  Systems (IROS)}, pages 3012--3018. ieee, 2007.

\bibitem{okada2011jfr}
Y.~Okada, K.~Nagatani, K.~Yoshida, S.~Tadokoro, T.~Yoshida, and E.~Koyanagi.
\newblock Shared autonomy system for tracked vehicles on rough terrain based on
  continuous three-dimensional terrain scanning.
\newblock {\em Journal of Field Robotics (JFR)}, 28(6):875--893, 2011.

\bibitem{paolo2017arxiv}
G.~Paolo, L.~Tai, and M.~Liu.
\newblock Towards continuous control of flippers for a multi-terrain robot
  using deep reinforcement learning.
\newblock {\em arXiv preprint arXiv:1709.08430}, 2017.

\bibitem{pecka2017iros}
M.~Pecka, K.~Zimmermann, and T.~Svoboda.
\newblock Fast simulation of vehicles with non-deformable tracks.
\newblock In {\em Proc.~of the IEEE/RSJ Intl.~Conf.~on Intelligent Robots and
  Systems (IROS)}, volume 2017, pages 6414--6419, 2017.

\bibitem{silver2021ai}
D.~Silver, S.~Singh, D.~Precup, and R.S. Sutton.
\newblock Reward is enough.
\newblock {\em Artificial Intelligence}, 299, 2021.

\bibitem{suzuki2014robomech}
S.~Suzuki, S.~Hasegawa, and M.~Okugawa.
\newblock Remote control system of disaster response robot with passive
  sub-crawlers considering falling down avoidance.
\newblock {\em ROBOMECH Journal}, 1(1), 2014.

\bibitem{Wang2016icml}
Z.~Wang, T.~Schaul, M.~Hessel, H.~Van~Hasselt, M.~Lanctot, and N.~De~Frcitas.
\newblock Dueling network architectures for deep reinforcement learning.
\newblock In {\em Proc.~of the Int.~Conf.~on Machine Learning (ICML)},
  volume~4, pages 2939--2947, 2016.

\bibitem{zhang2014rss}
J.~Zhang and S.~Singh.
\newblock {LOAM: Lidar Odometry and Mapping in Real-time}.
\newblock In {\em Proc.~of Robotics: Science and Systems (RSS)}, 2014.

\bibitem{zimmermann2014icra}
K.~Zimmermann, P.~Zuzanek, M.~Reinstein, and V.~Hlavac.
\newblock Adaptive traversability of unknown complex terrain with obstacles for
  mobile robots.
\newblock In {\em Proc.~of the IEEE Intl.~Conf.~on Robotics \& Automation
  (ICRA)}, pages 5177--5182, 2014.

\end{thebibliography}
